\documentclass[final]{main}
\pdfoutput=1
\titre{Modèles de Substitution pour les Modèles à base d'Agents : Enjeux, Méthodes et Applications}

\auteur{Paul Saves\up{a}}{paul.saves@ut-capitole.fr} \auteur{Nicolas Verstaevel\up{a}}{nicolas.verstaevel@ut-capitole.fr}
\auteur{Benoît Gaudou\up{a}}{benoit.gaudou@ut-capitole.fr}

\institution{\up{a}%
  Univ. Toulouse Capitole, IRIT, Toulouse, France}

\usepackage{cuted}
\usepackage[super]{nth}
\usepackage{capt-of}
\usepackage{stfloats}
\usepackage{float}
 \usepackage{booktabs}
\usepackage{graphicx} 
\usepackage{colortbl}
\begin{document}
\maketitle
\begin{resume}
La simulation multi-agent permet de modéliser et d'analyser les comportements dynamiques ainsi que les interactions d'entités autonomes évoluant dans des environnements complexes. Toutefois, sa mise en œuvre peut être freinée par des coûts computationnels élevés, limitant son applicabilité à grande échelle. Les modèles de substitution offrent une alternative prometteuse en remplaçant ces simulations coûteuses par des approximations plus légères et efficaces.
Cet article explore les motivations et les différentes approches permettant d’approximer les simulations agents, en s’appuyant sur l’apprentissage machine, la quantification des incertitudes et les stratégies d’analyse de sensibilité explicables.
À travers une étude de cas, nous mettons en évidence les défis liés à la fiabilité et à l’interprétabilité de ces modèles, en comparant leurs performances et leurs limites. Enfin, nous proposons des perspectives pour l’intégration d’outils d’apprentissage machine dans divers domaines.
\end{resume}
\motscles{Modèles de substitution, Modèles à base d'agents, Simulation, Analyse boîte noire}
\bigskip
\begin{abstract}
Agent-based models are powerful tools for simulating complex systems with autonomous entities interacting in dynamic environments. However, their high computational costs limit large-scale applications. Surrogate models offer a promising solution by replacing costly simulations with faster, efficient approximations. This article explores the motivations, methods, and applications of surrogate models for agent models, focusing on machine learning techniques, uncertainty quantification, and optimization strategies. Through the study of a practical test case, we present the challenges related to the reliability and interpretability of these models and compare their performance and limitations. We thus propose perspectives for the integration of machine learning tools in various fields.
\end{abstract}
%
%
\section{Introduction}
Les modèles à base d'agents (ABM) sont des outils puissants pour simuler des systèmes complexes composés d'agents autonomes qui interagissent entre eux et avec leur environnement. Ces modèles sont particulièrement important pour étudier les phénomènes émergents, par exemple lorsque des interactions locales donnent lieu à des comportements globaux, souvent difficiles à capturer avec des méthodes analytiques traditionnelles~\cite{debosscher2023}.
Cependant, les ABM présentent des défis computationnels importants. À mesure que le nombre d'agents et d'interactions augmente, le coût computationnel des simulations croît de manière exponentielle. Cela limite leur évolutivité et leur applicabilité à des simulations à grande échelle~\cite{gaudou2014}. Par conséquent, la réalisation de simulations étendues pour des tâches telles que l'analyse d'incertitude ou l'optimisation devient souvent irréalisable en raison de l'importance des temps de calculs.
Pour surmonter ces limitations, les modèles de substitution constituent une solution prometteuse offrant des approximations computationnellement efficaces~\cite{llacay2025}. Ces modèles apprennent directement sur les données issues de  simulations préalables et permettent de réduire considérablement le coût de la simulation des ABM en donnant des approximations des réponses attendues quasi instantanément. Les modèles de substitution permettent ainsi une analyse à grande échelle et une prise de décision en temps réel et rendent possible l'exploration de scenarios autrement inaccessibles en raison des contraintes computationnelles~\cite{angione2022}.

Dans cet article, nous proposons une analyse approfondie des enjeux, des méthodes et des applications des modèles de substitution pour les simulations multi-agents. Notre contribution se décline en deux volets principaux :
\begin{itemize}
\item    Une analyse comparative des méthodes d’apprentissage permettant de construire des modèles de substitution robustes, capables de quantifier l’incertitude et d’offrir une interprétabilité fine des prédictions.
    \item Une application concrète au modèle de ségrégation, illustrant les défis liés à la construction, à la validation et à l’interprétation des modèles.
 
\end{itemize} 
\vspace{0.1cm} Par rapport à l’état de l’art, nous mettons un accent particulier sur la quantification de l'incertitude, de la stochasticité et de l’explicabilité des modèles. 
L’objectif n’est pas seulement d’obtenir un modèle performant, mais aussi d’améliorer la confiance dans ses prédictions en évaluant la sensibilité aux variations des paramètres. 
Les codes et données ayant servis à produire les résultats sont disponibles en ligne\footnote{\url{https://github.com/ANR-MIMICO/JFSMA_Schelling}}.
Dans la Section~\ref{sec:sota}, nous présenterons les fondements des ABM et les défis qu’ils soulèvent, puis nous introduirons les modèles de substitution comme solution pour accélérer les simulations complexes.  Ensuite, la Section~\ref{sec:app} présentera une application des modèles de substitution pour une simulation multi-agent d'un  problème de ségrégation. Pour terminer, l'article se conclura en Section~\ref{sec:conclu} sur une réflexion générale sur les perspectives de recherche et les défis à relever pour améliorer la fidélité des modèles et les méthodes explicables dans le cadre des simulations complexes. 
\section{Simuler des systèmes complexes}
\label{sec:sota}
Les simulations de systèmes complexes s’appuient sur une combinaison de données expérimentales, de connaissances physiques et de mécanismes d’interaction avancés. Elles permettent de reproduire des phénomènes dynamiques en intégrant ces différentes sources d’information, mais peuvent s’avérer coûteuses en temps de calcul. Cette section présente les fondements des modèles à base d'agents ainsi que les modèles de substitution, qui visent à accélérer les simulations de  systèmes physiques.
\subsection{Les Modèles à base d'Agents : Fondations et Défis}
Les ABM offrent un cadre computationnel puissant pour simuler des systèmes complexes composés d'agents autonomes interagissant dans un environnement dynamique. Ces modèles ont été largement utilisés dans divers domaines tels que l'épidémiologie,  l'économie ou l'aménagement urbain, fournissant une approche ascendante pour comprendre les phénomènes émergents~\cite{treuil2008}.
Un modèle à base d'agents se compose généralement de trois éléments fondamentaux : (i) des \textit{agents}, qui opèrent selon des règles comportementales prédéfinies ou adaptatives ; (ii) un \textit{environnement}, qui structure les interactions spatiales et en réseau ; et (iii) des \textit{mécanismes d'interaction}, qui régissent le comportement des agents et l'évolution du système~\cite{treuil2008}. 
De plus, les simulations multi-agents sont généralement stochastiques car elles intègrent des éléments probabilistes, notamment dans le processus de prise de décision des agents. 

Néanmoins, malgré leurs avantages théoriques, les ABM rencontrent d'importants {défis computationnels} qui limitent leur évolutivité et leur utilisation en temps réel. À mesure que le nombre d'agents et d'interactions augmente, le coût computationnel s'accroît de façon exponentielle, rendant les simulations à grande échelle souvent irréalisables. De plus, ces modèles requièrent une {calibration des paramètres} rigoureuse, impliquant des milliers de simulations pour obtenir un ajustement fiable, et l'{identification des incertitudes} demeure problématique, puisque de légères variations dans les paramètres d'entrée peuvent entraîner des divergences significatives dans les résultats. Face à ces limitations, il est nécessaire de recourir à des techniques de modélisation de substitution qui permettent d'approximer les résultats des ABM tout en conservant la dynamique essentielle du système~\cite{taillandier2019}.
\subsection{Les Modèles de Substitution pour accélérer les simulations}
Les modèles de substitution visent à offrir des approximations efficaces des résultats issus de simulations complexes, permettant une évaluation rapide de la dynamique du système tout en réduisant considérablement les coûts computationnels associés aux ABM à grande échelle~\cite{angione2022}. Ils se déclinent en plusieurs catégories~\cite{saves2024} : 
\begin{itemize}
  \item \textbf{Approximations analytiques} : qui proposent des représentations mathématiques simplifiées du comportement du système.
  \item \textbf{Modèles basés sur les données} : qui exploitent des techniques d'apprentissage machine pour établir des relations entre les paramètres d'entrée et les résultats de simulation.
  \item \textbf{Modèles d'ordre réduit} : qui extraient les motifs dominants du système pour réduire la dimensionnalité.
\end{itemize}

Les approches basées sur l'apprentissage machine, telles que les \textit{processus gaussiens (GP)}, permettent d'obtenir des approximations probabilistes tout en intégrant une quantification des incertitudes~\cite{saves2024}. 
De même, les {réseaux de neurones (NN)} et les \textit{méthodes d'apprentissage ensemblistes} (par exemple, les forêts aléatoires ou le gradient boosting) offrent des représentations flexibles et de haute dimension des comportements des ABM, permettent d'intégrer des informations contextuelles locales et peuvent également donner des informations sur l'incertitude~\cite{martin2024, blanco-volle2024}. 

Cependant, l'utilisation de modèles de substitution introduit des \textit{compromis} : bien qu'ils réduisent significativement les coûts de calcul, ils peuvent compromettre la précision du modèle, notamment lors de l'extrapolation au-delà des données d'apprentissage. Il est donc impératif de valider rigoureusement ces modèles pour garantir leur \textit{généralisabilité} dans des espaces de paramètres à haute dimension. De plus, l'\textit{interprétabilité} des modèles complexes basés sur l'apprentissage machine reste une préoccupation majeure, surtout dans les domaines nécessitant une transparence totale des résultats ou une forte confiance dans les modèles.
\subsection{Optimisation et validation de modèles}
L'optimisation et la validation des modèles de simulation complexes sont des étapes clés pour assurer la fiabilité et la précision des prédictions. D'une part, la calibration vise à ajuster les paramètres du modèle afin de reproduire fidèlement le comportement observé dans le système réel. D'autre part, des méthodes statistiques telles que l'optimisation bayésienne permettent d'identifier le jeu de paramètres optimal tout en limitant le nombre d'appels au modèle ABM, souvent traité comme une boîte noire coûteuse sur laquelle peu de données sont disponibles. 
Dans ce contexte, plusieurs travaux récents apportent des contributions complémentaires. Par exemple, il a été montré que pour la prédiction d'une surface de réponse, l'apprentissage machine peut servir de substitut aux simulations multi-agents~\cite{angione2022}. De plus, les méthodes ensemblistes peuvent améliorer l'interprétabilité des modèles de substitution, offrant ainsi une vue détaillée de l'impact de chaque variable sur la prédiction~\cite{blanco-volle2024}. Par ailleurs, des approches d'inférence bayésienne en boîte noire pour la calibration des ABM, permettent une estimation robuste des paramètres et des incertitudes malgré la complexité inhérente aux systèmes simulés~\cite{dyer2024}. De plus, des modèles spécialisés peuvent identifier les points de basculement dans les modèles multi-agents, ouvrant la voie à l'anticipation des transitions critiques dans des systèmes dynamiques~\cite{fabiani2024,blanco-volle2024}. 

Dans cet article, nous proposons d'aller plus loin en intégrant des questions de validation de l’explicabilité et de quantification de l'incertitude~\cite{robani2025}. L'utilisation combinée de ces approches permet d'identifier d'éventuelles insuffisances et d'apporter les ajustements nécessaires pour améliorer la confiance en la performance globale des modèles. 
%
\section{Application au modèle de ségrégation}
\label{sec:app}
%
%
%
Le modèle de ségrégation résidentielle de \textit{Thomas Schelling} constitue une étude classique des effets des décisions locales sur les dynamiques globales. Il met en évidence comment des agents possédant de légères préférences pour des voisins de même type, sans intention explicite de créer des quartiers homogènes, peuvent néanmoins conduire à une ségrégation complète~\cite{schelling1969}. Une analyse globale des paramètres de ce modèle 
coûteux à évaluer ont mis en lumière des comportements émergents de différentes nature et plusieurs variables intéressantes à explorer pour un modèle de substitution voulant apprendre sur peu de données~\cite{singh2009}.
\begin{figure*}[htb]
\centering
\includegraphics[height=4.0cm,width=0.75\textwidth]{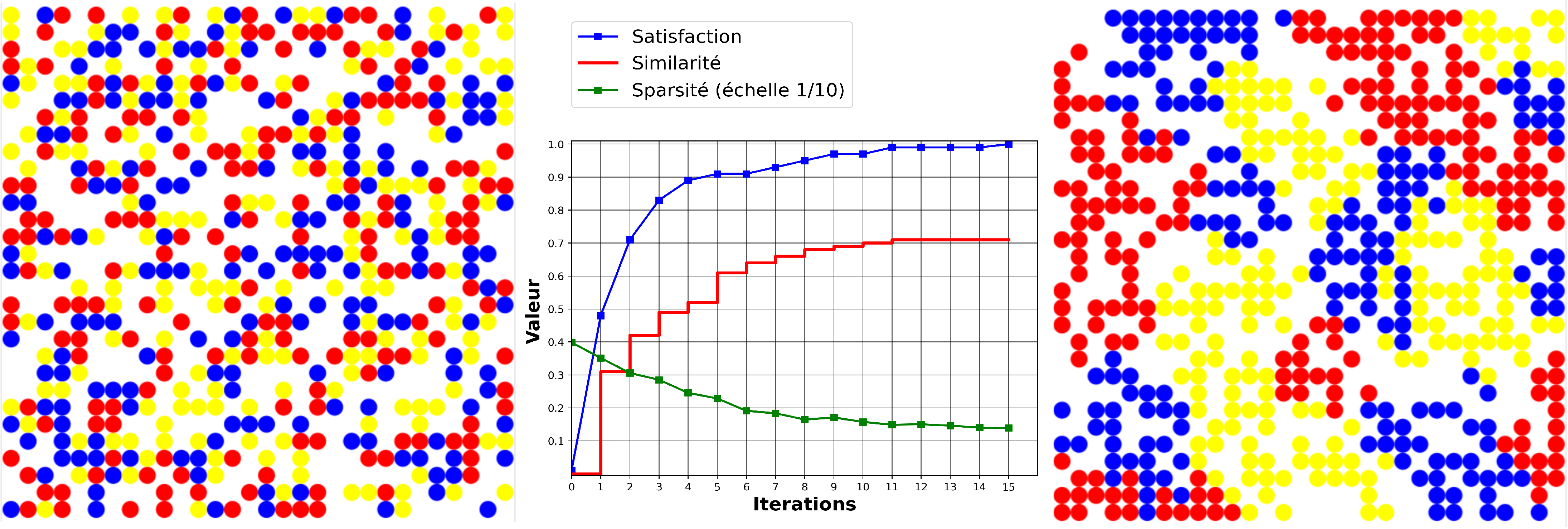}
\caption{Un exemple de simulation du problème de séggregation avec la plateforme GAMA}
\label{fig:gamma_simu}
\end{figure*}
\subsection{Présentation du modèle simulé}
Dans ce modèle, les agents sont initialement placés de manière aléatoire sur une grille, avec une certaine \textit{densité} et se voit attribué un \textit{type} (représenté par une couleur). Leur état est déterminé en fonction de leur environnement immédiat et peut être soit \textbf{Satisfait} (happy), lorsque la proportion de voisins d’un type différent est inférieure à leur \textit{seuil d'intolérance maximal} soit  \textbf{Insatisfait} (unhappy) dans le cas contraire.
Lorsqu’un agent est insatisfait, il se déplace aléatoirement 
sur une case libre.
En complément des paramètres précédents, il est possible d’ajuster la \textit{distance de perception} des agents, définie comme la portée à laquelle ils considèrent les autres comme voisins. 

La simulation repose principalement sur les cinq variables suivantes :  
\begin{itemize}
    \item Le nombre entier de types d’individus considérés entre 2 et 5.
    \item La densité d’individus entre 0.01 et 1.
    \item Le seuil d’intolérance entre 0 et 1.  
    \item La taille de la carte carrée de $10 \times 10$ à $40 \times 40$.
    \item La distance entière de perception de $1$ à $10$ en distance euclidienne.
\end{itemize}

Il existe également des sources d'incertitude aléatoire exogènes qui viennent de la position initiale des individus et leurs déplacements d’une itération à l’autre.
Il est à noter qu'une simulation peut ne jamais converger car les paramètres empêchent un équilibre d'exister. Plus surprenamment, un équilibre peut exister sans être nécessairement atteint en temps fini, du fait que tous les individus se déplacent de manière aléatoire et indépendamment de leur passé. 

Sur le plan de l'implémentation, nous utiliserons le modèle de ségrégation  présent dans la librairie de modèles de la plateforme GAMA~\cite{taillandier2019}. 
%
La Figure~\ref{fig:gamma_simu} illustre un exemple de simulation avec trois types d'agents, une densité de 60\%, un seuil d'intolérance de 33\%, une distance de perception de 3 unités et une grille de $30 \times 30$. Les cases sans agents restent vides, tandis que les agents sont colorés selon leur type.   
À gauche, l’état initial est généré aléatoirement, avec un pourcentage moyen d'individus présents dans le voisinage des agents, appelé similarité, quasiment nul et de nombreux agents insatisfaits. Progressivement, le taux d'agents satisfaits — qui cessent alors de se déplacer — augmente jusqu'à atteindre un état stable après 15 itérations. Cet état stable, représenté sur la carte de droite, correspond à une dissimilarité moyenne de 30\%.  La sparsité qui caractérise à quel point les agents sont éparpillés, diminue progressivement à mesure que les agents se regroupent jusqu'à se stabiliser vers 1.4. 
\subsection{Génération des données}
Afin d'explorer l'espace des paramètres de manière efficace, nous avons généré un échantillon par hypercubes latin imbriqués, basé sur une distribution uniforme, selon l’hypothèse de criblage~\cite{arenzana2021}. Nous avons généré un plan d’expérience de 200 points, structuré de manière à contenir des sous-ensembles de 100 et 50 points également hypercubes latin.
Les hypercubes latins ont été générés en suivant le critère ESE (Enhanced Stochastic Evolutionary algorithm) tel qu'implémenté dans la toolbox SMT (Surrogate Modeling Toolbox)~\cite{jin2005}. De plus, pour prendre en compte la variabilité inhérente aux aspects aléatoires du système, nous avons fait cinq répétitions par point du plan d’expérience. Cela représente un total de 1000 simulations ($200 \times 5$).
On obtient ainsi un jeu d'entraînement de 25\% et un jeu de validation de 75\% du jeu de données complet ce qui nous  permettra de tester si un apprentissage sur un petit jeu de données arrive à reproduire les analyses obtenues à partir d'une exploration plus exhaustive. 

Afin de valider notre approche à la fois pour la régression et pour la classification, nous considérons deux types de sorties : une variable binaire indiquant la convergence et une variable continue représentant la sparsité~\cite{singh2009}.
Une simulation s'arrête si plus aucun individu ne se déplace encore sur la grille. Nonobstant, et sachant qu'une simulation peut ne jamais converger, nous avons fixé un seuil de 1000 itérations, au-delà duquel la simulation est arrêtée et considérée comme non convergente. 
Le rapport entre le nombre de voisins qui ne sont pas semblables et le nombre de voisins similaires est appelé mesure \(u/l\). Pour un agent sur la case de coordonnées \((i, j)\), le rapport \(u/l\) de l'agent est donné par :
\begin{equation}
[u/l]_{i,j} = \frac{q_{i,j} + w_{i,j}}{s_{i,j}}
\end{equation}
où \(s_{i,j}\), \(q_{i,j}\) et \(w_{i,j}\) désignent respectivement le nombre de voisins similaires, différents et les cases vacantes entourant l’agent, considéré voisin de lui même pour éviter de diviser par $0$. La sparsité (ou parcimonie) \(\langle [u/l] \rangle\) d’un ensemble d’agents est obtenue en moyennant la mesure \([u/l]\) sur l’ensemble des agents :

\begin{equation}
\langle [u/l] \rangle = \frac{1}{N} \sum_{(i,j) \in C} [u/l]_{i,j}
\end{equation}

où \(N\) est le nombre total d’agents dont les cases associées forment l'ensemble \(C\). Pour une carte de côté $c$2 et une densité $\rho$, on a $N = \lfloor \rho c^2 \rfloor $.

Une simulation dure de 3 secondes à 30 minutes sur un processeur Intel Core Ultra 9 185H vPro. On peut donc considérer le modèle agent comme une boîte noire coûteuse ce qui motive l'usage de modèles de substitution dans ce contexte.
\subsection{Analyse du jeu de données}
\begin{figure*}[htb]
\centering
\includegraphics[height=6cm, width=0.75\textwidth]{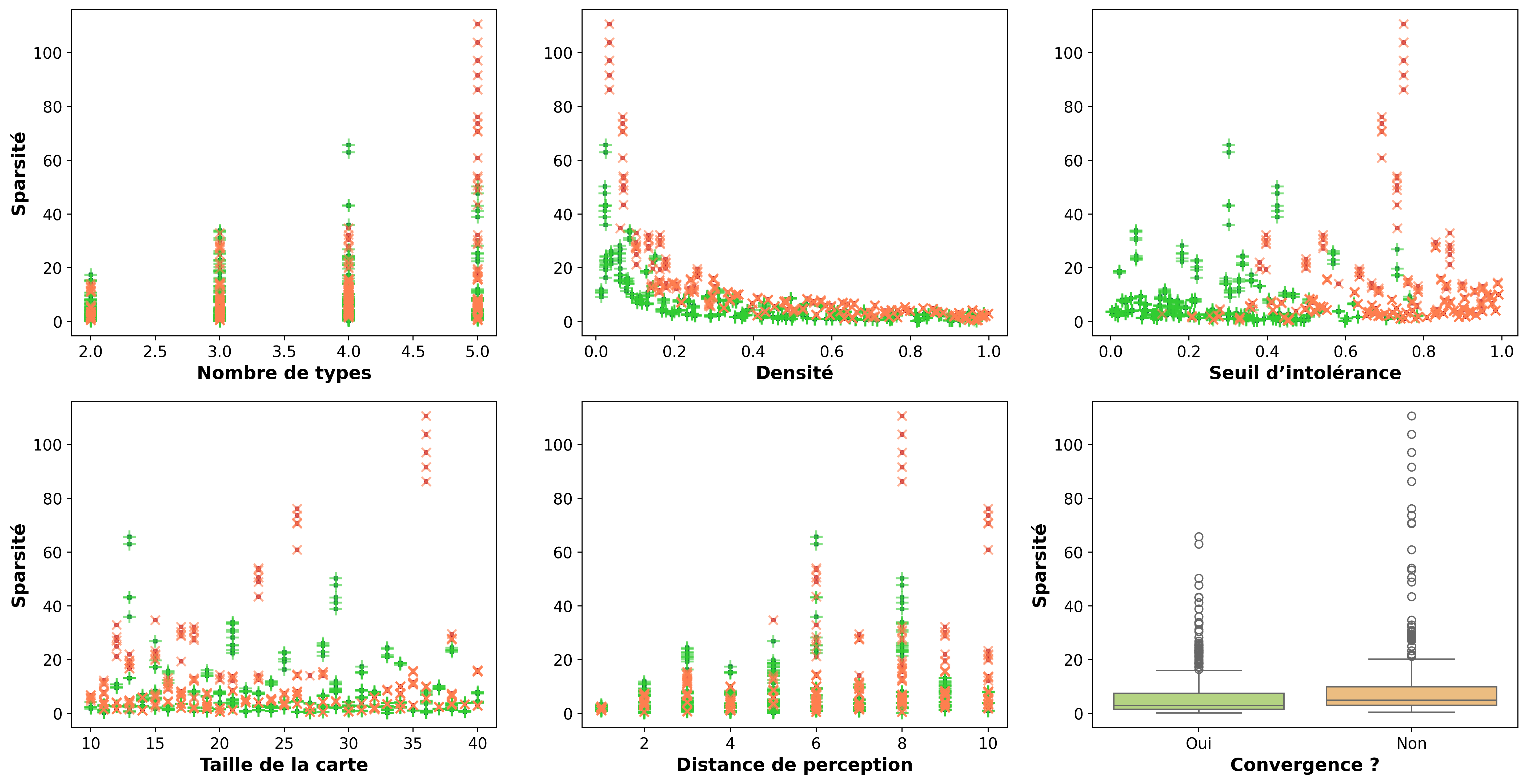}
\caption{Le jeu de donnée avec en vert les simulations convergentes et en rouge les autres}
\label{fig:data_seg}
\end{figure*}

Le plan d'expérience complet est illustré sur la Figure~\ref{fig:data_seg} avec, pour chacune des 5 variables, les valeurs de l'indice de sparsité en fonction de la valeur de la variable. En vert sont tracées les simulations qui ont convergé tandis qu'en rouge figurent les simulations qui ont été interrompues après 1000 itérations. La distribution des valeurs de sparsité est donné sur la sixième figure sous forme de boîte à moustaches.

La sparsité reflète le niveau de mélange des agents dans la simulation. Une valeur élevée indique soit un fort mélange d’agents soit de nombreuses cases vides, traduisant ainsi une faible ségrégation et une entropie spatiale élevée. À l'inverse, une sparsité faible signifie que la majorité des agents sont entourés de voisins similaires, indiquant une forte ségrégation et une organisation spatiale plus ordonnée, ce qui se traduit par une entropie réduite. En analysant l’évolution temporelle de cette métrique, on peut rapidement anticiper la dynamique du système : après quelques dizaines d'itérations, les simulations convergentes montrent une décroissance rapide de la sparsité quand les agents se regroupent avant de finir sur un plateau plus stable une fois que les groupes sont formés, tandis que les simulations non convergentes présentent des fluctuations persistantes et oscillantes à cause des groupes qui se font et se défont. Cette distinction permettrait de prédire l’état final du système pour classifier et prédire le comportement des agents à partir de séries temporelles.

Sur 1000 simulations, 546 ont convergé et finissent avec un sparsité moyenne de 6.20 contre 9.27 pour les 454 simulations non convergés. A noter que sur les 200 expériences, 16 convergent partiellement, dépendamment de l'aléatoire sur les 5 répétitions. Ces 16 configurations partiellement convergentes se situent à des endroits précis de l'espace de recherche. Les valeurs d'intolérance se situent près de la limite de convergence, entre 0.4 et 0.8, le nombre de type est généralement de 3 ou 4, la taille de  carte médiane est de 18.5, et la distance de perception est généralement de 3 à 9. 
Cela est probablement du au fait que les  cartes avec peu de degrés de libertés sont plus dépendantes de la position initiale aléatoire. Les petites valeurs de distances de perception ou de nombre de types sont plus simple à converger et inversement, favorisant les consensus sur la décision. Pour finir, on obtient plus facilement des comportements aléatoires lorsque le seuil d'intolérance est proche de la valeur de bascule vers 0.6.

Une rapide analyse descriptive a porté sur la relation entre les caractéristiques d'entrée et les deux sorties. D'abord, concernant la corrélation de Pearson sur le jeu de données, on remarque que le nombre de types d'individus est corrélé à la sparsité (0,25), ce qui était attendu puisque plus il y a d'agents non semblables, plus la sparsité augmente, bien que cela diminue légèrement le taux de convergence (–0,18). La densité d'individus diminue significativement la sparsité (–0,56) et le taux de convergence (–0,19), car les agents ont moins de place et donc plus de voisins. Le seuil d'intolérance est légèrement corrélé avec la sparsité (0,10) mais empêche fortement la simulation de converger (–0,75). La taille de la carte n'a aucune corrélation significative avec la sparsité (0,004) et présente une faible corrélation positive avec la convergence (0,10). Enfin, la distance de perception a une corrélation positive modérée avec la sparsité (0,25) tout en diminuant le taux de convergence (–0,19) puisque des individus plus exigeants recherchent des quartiers plus séparés.
Des tests ANOVA confirment la significativité du nombre de types d'individus, de la densité d'individus et du seuil d'intolérance dans la détermination de la convergence. De plus, la \textit{p-value} indique que la densité d'individus et le seuil d'intolérance sont les variables les plus influentes sur la convergence, tandis que la taille de la carte semble de faible importance. Ces résultats suggèrent que la densité d'individus et le seuil d'intolérance sont des paramètres déterminants, que le nombre de types d'individus et la distance de perception jouent également un rôle mais que la taille de la carte n'influence que peu les sorties.
\subsection{Modèles de substitutions}
De nombreux travaux ont déjà montré l'intérêt des modèles de substitution pour apprendre des simulations à base d’agents pour la prédiction de surface de réponse~\cite{angione2022, fabiani2024,llacay2025}. Dans un premier temps, 
nous allons tester cette approche sur le modèle de Ségrégation en apprenant un modèle de substitution avec 50 des 200 points répétés 5 fois. 
Dans l’ensemble de nos expériences, nous utilisons les options par défaut des modèles en l’absence de modifications spécifiques. On a donc deux problèmes à résoudre, à savoir la prédiction par régression de la sparsité et la prédiction par classification de la convergence, à chaque fois avec des modèles d'interpolations entre des bornes fixées et connues.
Afin de valider nos modèles, nous avons établi un plan d’expérience comportant 200 points, ce qui correspond à 150 nouvelles configurations et génère un total de 750 nouvelles données. Sur l’ensemble de la base complète, nous évaluons la performance en calculant la racine de l'erreur quadratique moyenne (RMSE) pour la prédiction de la sparsité et en comptant le nombre d’erreurs de classification parmi les 750 nouvelles observations.




Concernant les modèles testés, le processus gaussien (GP) intègre un bruit hétéroscédastique, évalué automatiquement via les expériences répétées~\cite{arenzana2021} auquel est associé un terme de tendance linéaire (krigeage  ordinaire) et une fonction de corrélation exponentielle carrée pour la régression et exponentielle absolue pour la classification. Pareillement, les fonctions en base radiale (RBF) utilise une tendance linéaire et la méthode d’interpolation par pondération inverse de la distance (IDW) utilise avec une fonction de corrélation exponentielle carré pour la régression et une fonction exponentielle absolue pour la classification. Pour les modèles de régression linéaire (LR) ou polynomiale quadratique (QP), nous utilisons les options par défaut et pour le modèle de splines régularisées à produit tensoriel d'énergie minimale (RMTS), nous utilisons la formulation à base de splines cubiques d'Hermite. Enfin, pour le modèle de réseau de neurones amélioré par gradient (GENN), nous exploitons les dérivées prédites par le GP, enrichissant ainsi l’information utilisée lors de l’entraînement du réseau.

Nous comparons également ces modèles avec des méthodes issues de Scikit-learn~\cite{pedregosa2011}. Ainsi, pour la régression et la classification, nous utilisons notamment l’arbre de décision (DT), la forêt aléatoire (RF), les k plus proches voisins (kNN), le perceptron multicouche (MLP) ainsi que la machine à vecteurs de support (SVM). Pour la méthode kNN, nous avons choisi d’utiliser 15 voisins, correspondant à 3 emplacements différents à cause des répétitions.  Toutes ces méthodes viennent en deux versions, une version régression pour prédire la sparsité et une version classification pour prédire la convergence ou non de l'ABM. Comme pour scikit-learn, le Gradient Boosting de la librairie XGBoost traite distinctement la régression et la classification. 
Les résultats sont donnés dans la Table~\ref{tab:results}.
%
\begin{table}[ht]
    \centering
    \caption{Comparaison des performances des modèles avec classement}
    \resizebox{\linewidth}{!}{ 
    \begin{tabular}{lcccc}
        \toprule
            \textbf{Modèle} & \textbf{RMSE} & \textbf{Class. err.} & \textbf{Rg Rég.} & \textbf{Rg Class.} \\
        \midrule
        GP   & {\cellcolor[HTML]{05712F}} \color[HTML]{F1F1F1} 1.81  & {\cellcolor[HTML]{2C944C}} \color[HTML]{F1F1F1} 79   & 2  & 2  \\
        RF   & {\cellcolor[HTML]{00441B}} \color[HTML]{F1F1F1} 1.69  & {\cellcolor[HTML]{2C944C}} \color[HTML]{F1F1F1} 86   & 1  & 3  \\
        RBF  & {\cellcolor[HTML]{8ED08B}} \color[HTML]{000000} 1.95  & {\cellcolor[HTML]{00441B}} \color[HTML]{F1F1F1} 64   & 6  & 1  \\
        GB   & {\cellcolor[HTML]{56B567}} \color[HTML]{F1F1F1} 1.87  & {\cellcolor[HTML]{56B567}} \color[HTML]{F1F1F1} 91   & 4  & 5  \\
        DT   & {\cellcolor[HTML]{2C944C}} \color[HTML]{F1F1F1} 1.86  & {\cellcolor[HTML]{8ED08B}} \color[HTML]{000000} 123  & 3  & 7  \\
        LR   & {\cellcolor[HTML]{BCE4B5}} \color[HTML]{000000} 2.13  & {\cellcolor[HTML]{56B567}} \color[HTML]{F1F1F1} 90   & 7  & 4  \\
        QP   & {\cellcolor[HTML]{2C944C}} \color[HTML]{F1F1F1} 1.88  & {\cellcolor[HTML]{8ED08B}} \color[HTML]{000000} 129  & 5  & 8  \\
        RMTS & {\cellcolor[HTML]{BCE4B5}} \color[HTML]{000000} 2.19  & {\cellcolor[HTML]{BCE4B5}} \color[HTML]{000000} 260  & 8  & 10 \\
        IDW  & {\cellcolor[HTML]{E1F3DC}} \color[HTML]{000000} 2.26  & {\cellcolor[HTML]{BCE4B5}} \color[HTML]{000000} 130  & 9  & 9  \\
        MLP  & {\cellcolor[HTML]{DDF2D8}} \color[HTML]{000000} 2.52  & {\cellcolor[HTML]{8ED08B}} \color[HTML]{000000} 118  & 12 & 6  \\
        SVM  & {\cellcolor[HTML]{F7FCF5}} \color[HTML]{000000} 2.32  & {\cellcolor[HTML]{F7FCF5}} \color[HTML]{000000} 384  & 10 & 12 \\
        GENN & {\cellcolor[HTML]{E7F7E7}} \color[HTML]{000000} 2.37  & {\cellcolor[HTML]{F7FCF5}} \color[HTML]{000000} 384  & 11 & 12 \\
        kNN  & {\cellcolor[HTML]{F7FCF5}} \color[HTML]{000000} 2.57  & {\cellcolor[HTML]{E1F3DC}} \color[HTML]{000000} 345  & 13 & 11 \\
        \bottomrule
    \end{tabular}
    }
    \label{tab:results}
\end{table}

Nos résultats confirment que, pour résoudre les deux problématiques – la prédiction par régression de la sparsité et la prédiction par classification de la convergence – différents modèles de substitution présentent des performances variables et complémentaires.  

Pour la prédiction de la sparsité, le modèle basé sur la forêt aléatoire se démarque en affichant la meilleure performance avec une RMSE de 1.69, tandis que le modèle GP présente une performance comparable avec une RMSE de 1.81. Concernant l’erreur de classification, le modèle RBF se distingue particulièrement avec seulement 64 erreurs (8.5\%), ce qui en fait le modèle le plus performant sur ce critère, suivi du modèle GP (79 erreurs). Sur ce type de données parcimonieuses et uniformément réparties, il est généralement préférable d’utiliser un GP ou un RBF, car ces modèles apprennent rapidement et capturent efficacement les comportements non linéaires sur des petites bases de données~\cite{dyer2024}.
La forêt aléatoire, en tant qu’approche ensembliste, fournît ici des résultats comparables, ce qui s’explique par leur aptitude à approximativement reproduire les comportements des modèles à base d’agents. Le gradient boosting et l’arbre de décision  qui arrivent un peu derrière confirment aussi la capacité de ces méthodes à gérer les interactions complexes entre agents~\cite{singh2009}. Les méthodes ensemblistes présentent le désavantage de nécessiter un grand nombre de prédictions pour estimer les incertitudes, générant ainsi des surfaces de réponse très discontinues, ce qui peut, en revanche limiter leur capacité d’analyse, leur explicabilité et la prédiction de gradients mais ouvre la voie à de nombreuses pistes de recherches. 

Comme attendu, le modèle de Schelling est suffisemment complexe pour que la régression linéaire ne parvienne pas à capter toutes les nuances du modèles. D'autres approches telles que l’IDW, le MLP et surtout les méthodes kNN et SVM affichent des performances moins convaincantes, avec des écarts plus importants tant en régression qu’en classification. kNN et IDW souffrent des variables non significatives comme la taille de la carte ainsi que des expériences répétées, qui faussent la pertinence de la distance pour la prédiction même si la pondération aide IDW qui s'en sort légérement mieux.
Enfin, nos résultats suggèrent que l’utilisation de modèles hybrides comme avec le modèle GENN, qui intègre les dérivées prédites par un GP pour enrichir l’apprentissage du réseau de neurones, obtient de meilleures performances en régression. Néanmoins, cela dégrade les performances en classification puisque les gradients ne sont ni adaptés ni précis pour des sorties binaires.

\begin{figure}[H]
\centering
\includegraphics[height=3.8cm, width=0.35\textwidth]{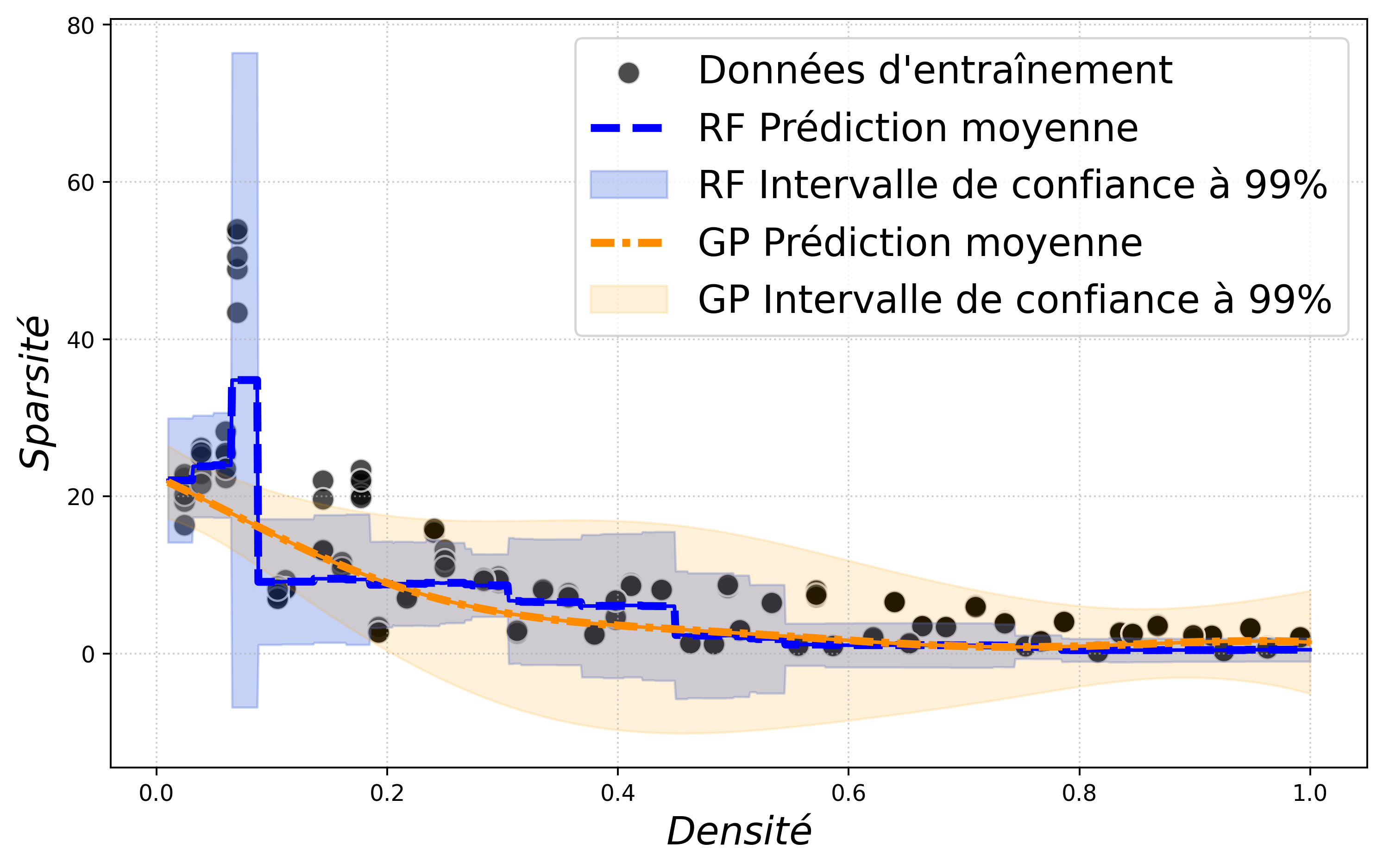}
\caption{Sparsité en fonction de la densité}
\label{fig:modeles_1D}
\end{figure}

Ces observations suggèrent que, dans notre étude, les modèles basés sur les processus gaussiens, les forêts aléatoires et les RBF offrent un compromis optimal entre la précision en régression et la performance en classification. Ces résultats fournissent ainsi une base solide pour une sélection judicieuse du modèle en fonction des applications ciblées.  Ces constatations appuient les récentes avancées dans l’utilisation de modèles de substitution pour les simulations à base d’agents, qui permettent non seulement d’accélérer les calculs, mais également d’obtenir des estimations d’incertitude cruciales pour l’interprétation des phénomènes émergents dans des systèmes complexes~\cite{fabiani2024}. Une comparaison entre le GP et la RF est donnée en 
Figure~\ref{fig:modeles_1D},  où chaque variable est fixée à la valeur moyenne (ou à sa partie entière dans le cas discret) sauf la densité qui varie entre $0$ et $1$, pour illustrer l'apect lisse du GP et l'aspect continu par morceaux de la RF et la mauvaise estimation de l'incertitude qui y est associée, obtenue par rééchantillonnage sur 100 estimateurs (options par défaut), là où le GP peut être directiment utilisé pour l'interprétabilité basée sur la variance~\cite{iooss2022}. 
\subsection{Analyses basées modèles}
\begin{figure}[b]
\centering
\includegraphics[height=3.5cm, width=0.26\textwidth]{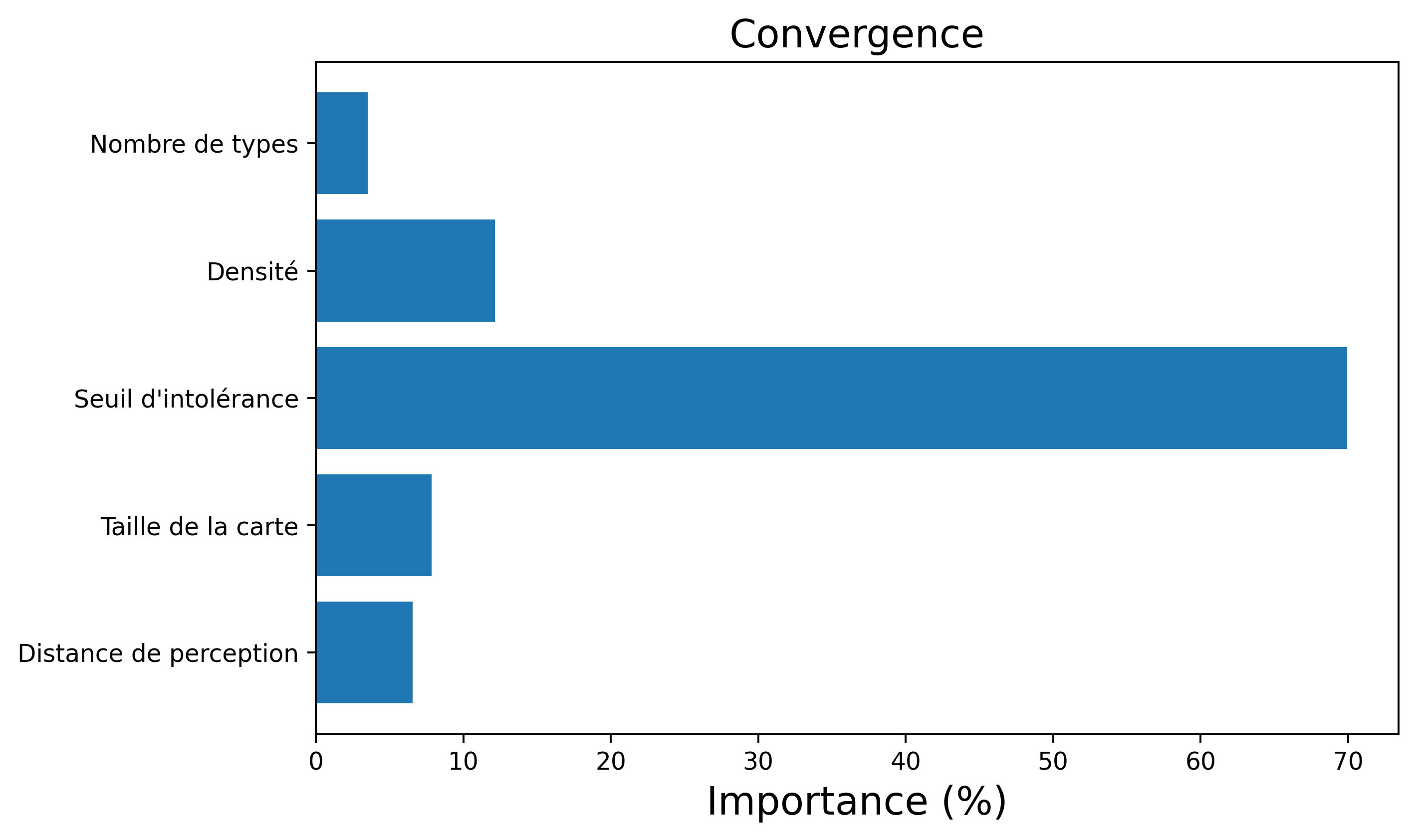}
\hspace{-0.25cm}
\includegraphics[height=3.5cm, width=0.21\textwidth]{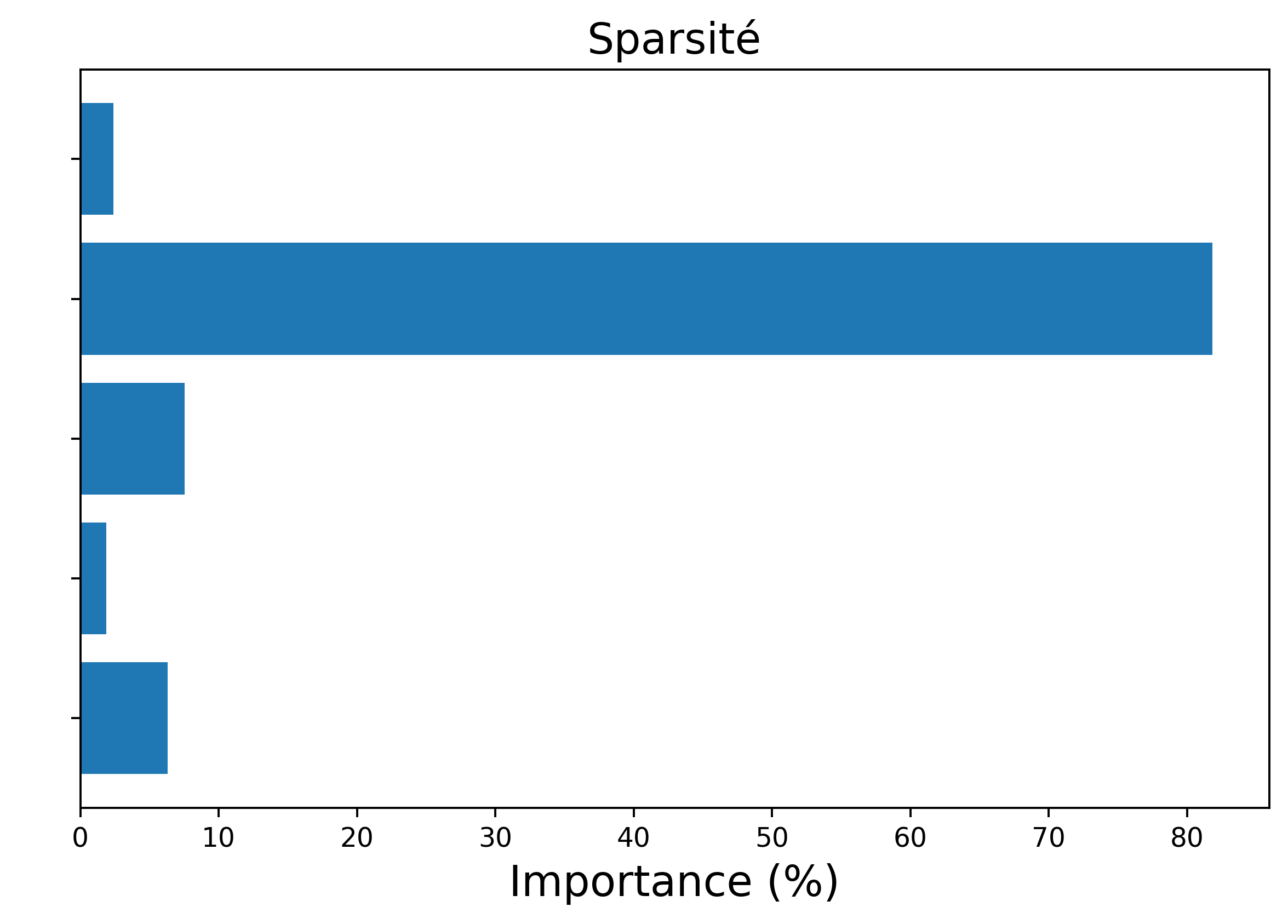}
\caption{Importance des variables dans la forêt}
\label{fig:imp_rf}
\end{figure}

Bien qu'étant limité et discontinus, les arbres de décisions sont des suites de procédures simple à interpréter. Par exemple, le premier nœud du GB ou du DT est "seuil d'intolérance inférieur à 0.63" ce qui confirme ce que l'on avait graphiquement sur le jeu de donnée complet.  
Dans une forêt aléatoire, l'importance d'une variable est souvent mesurée via la décroissance moyenne en impureté (MDI). À chaque division d'un arbre, on calcule la réduction de l'impureté, souvent mesurée par l'indice de Gini, qui quantifie la probabilité qu'un échantillon soit mal classé. Cette réduction est attribuée à la variable qui a permis le split, et en cumulant ces réductions sur tous les arbres, on obtient une mesure globale de l'importance de chaque caractéristique. Toutefois, cette approche peut biaiser l'estimation en faveur des variables à forte cardinalité, qui offrent davantage de possibilités de division. On obtient, de cette façon, sur notre cas test, les importances données en Figure~\ref{fig:imp_rf}.  Pour les deux sorties, on retrouve bien les variables qui influencent le plus les prédictions du modèle comme identifiées sur le jeu de donnée étendu, à savoir l'intolérance pour la convergence et la densité pour la sparsité. On retrouve également que la taille de la carte a une influence assez faible sur les sorties.

\begin{figure}[t]
\centering
\includegraphics[height=3.2cm, width=0.24\textwidth]{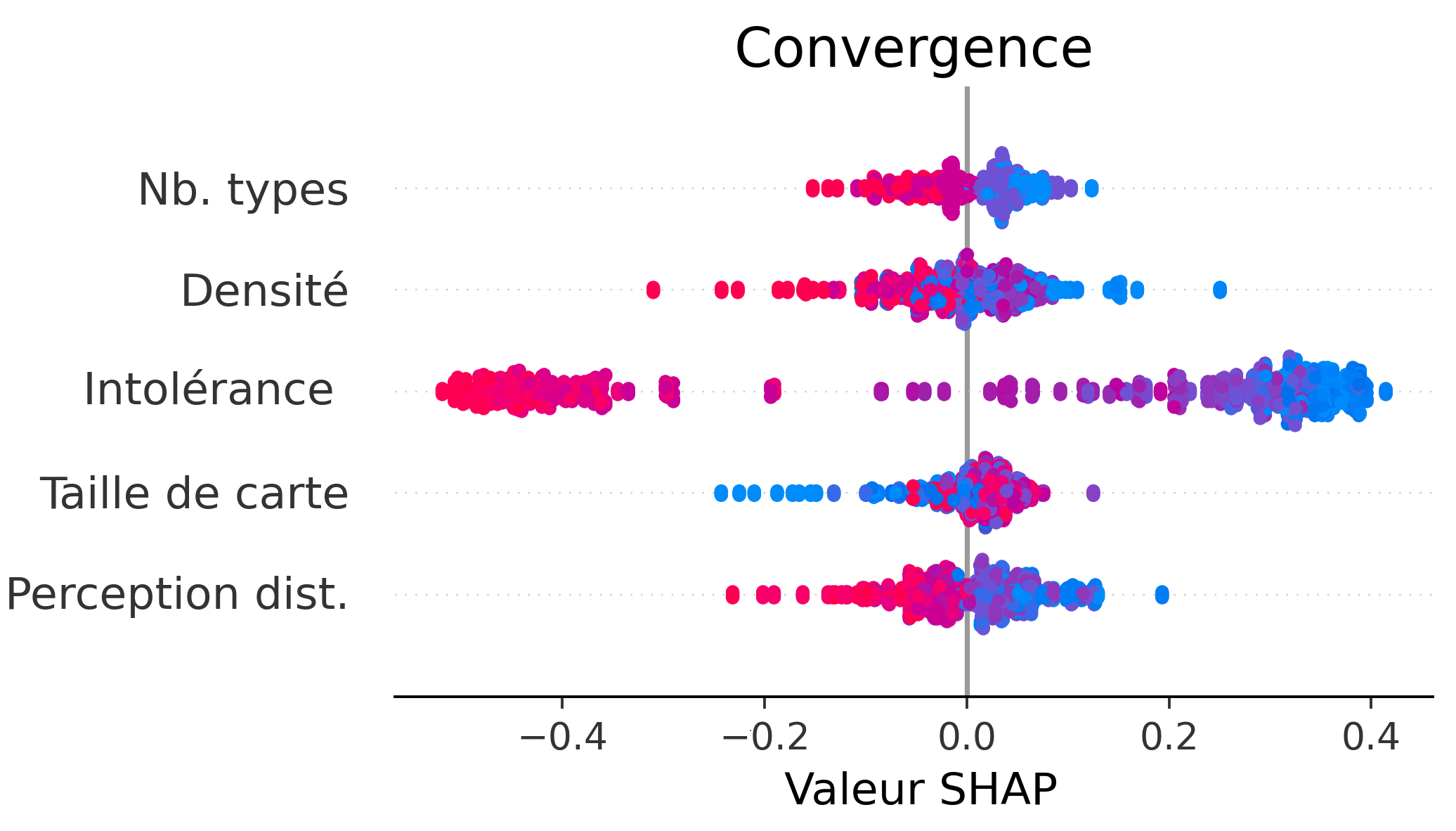}
\hspace{-0.25cm}
\includegraphics[height=3.2cm, width=0.23\textwidth]{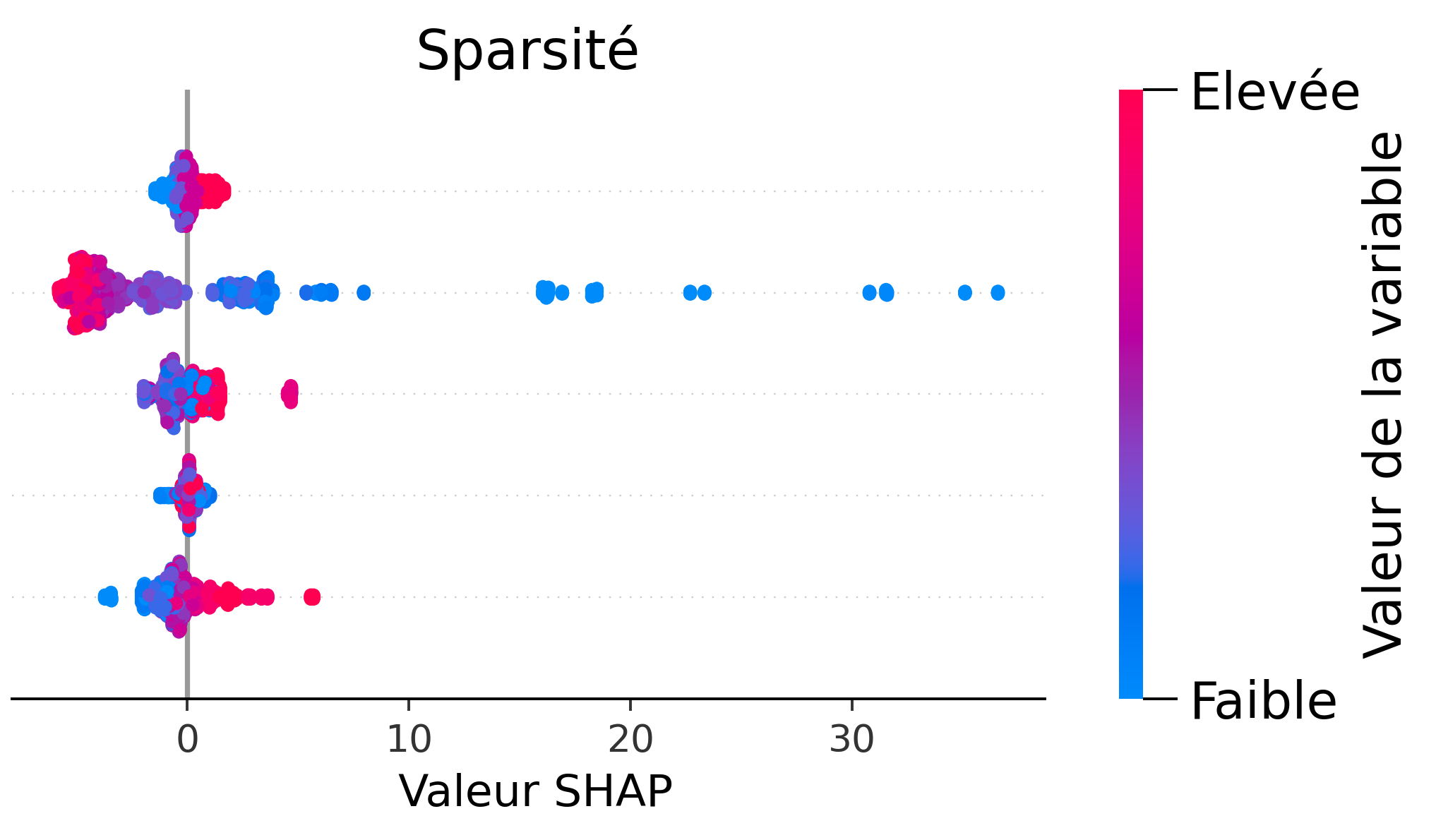}
\caption{Valeurs SHAP prédites par la forêt}
\label{fig:shap_rf}
\end{figure}

En complément, les valeurs SHAP (SHapley Additive exPlanations) fournissent une interprétation locale et fine quantifiant la contribution de chaque variable à la prédiction~\cite{shapley1953}. Ainsi, alors que la MDI offre une importance globale basée sur la somme des réductions d'impureté, les valeurs SHAP révèlent comment chaque caractéristique influence localement la sortie du modèle, permettant une analyse plus robuste et nuancée des prédictions. Les valeurs SHAP sont calculées à partir des prédictions du modèle sur le jeu de données de 1000 points et tracées en Figure~\ref{fig:shap_rf}, pour un coût de calcul très faible, les modèles de substitution étant rapides à évaluer.
Les valeurs SHAP permettent d’évaluer précisément la contribution de chaque paramètre et la distribution de ces contributions. On observe par exemple que l’intolérance et la densité ont un impact majeur, ce qui se traduit par des variations importantes dans la convergence ou la structure spatiale. Ainsi, l’analyse par SHAP complète l’interprétation globale de la forêt en montrant localement comment chaque paramètre oriente la prédiction. Par exemple, l'analyse révèle que l'intolérance a un pouvoir explicatif extrêmement fort, relativement centré et symétrique sur la convergence de la simulation. Au contraire,  pour les faibles valeurs de densité, la sparsité explose et écrase les autres variables, révélant une explicabilité plus locale de la densité sur la sparsité.
On peut faire les mêmes analyses d'importance des variables et de valeurs SHAP avec les prédictions du GP au lieu de celles de la RF. Pour changer, on utilise KernelSHAP, une version de SHAP qui fonctionne de manière similaire à LIME en générant des exemples perturbés et en ajustant un modèle linéaire et on calcule l'importance des variables à partir des valeurs SHAP prédites. On obtient ainsi des résultats globalement similaires à ceux obtenus sur la forêt, surtout pour l'importance des variables. Une différence notable est que le GP identifie, comme on pouvait s'y attendre, que certaines configurations à 5 types d'individus influencent significativement la sparsité, ce qui augmente l'influence du nombre de types : les individus ont moins de voisins similaires si la simulation présente une plus grande proportion d'individus différents.

\begin{figure}[t]
\centering
\includegraphics[height=4.25cm, width=0.47\textwidth]{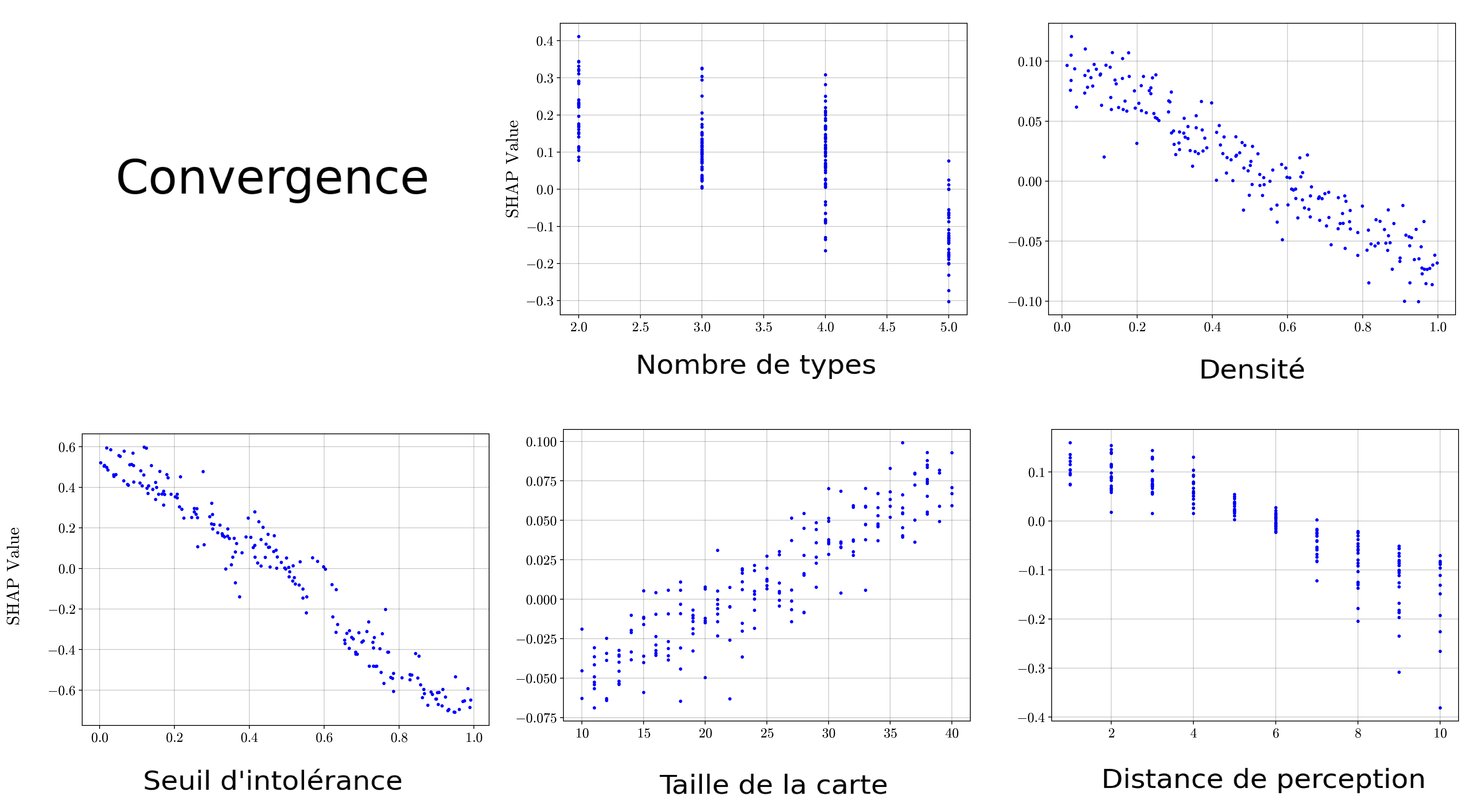}
\includegraphics[height=4.25cm,  width=0.47\textwidth]{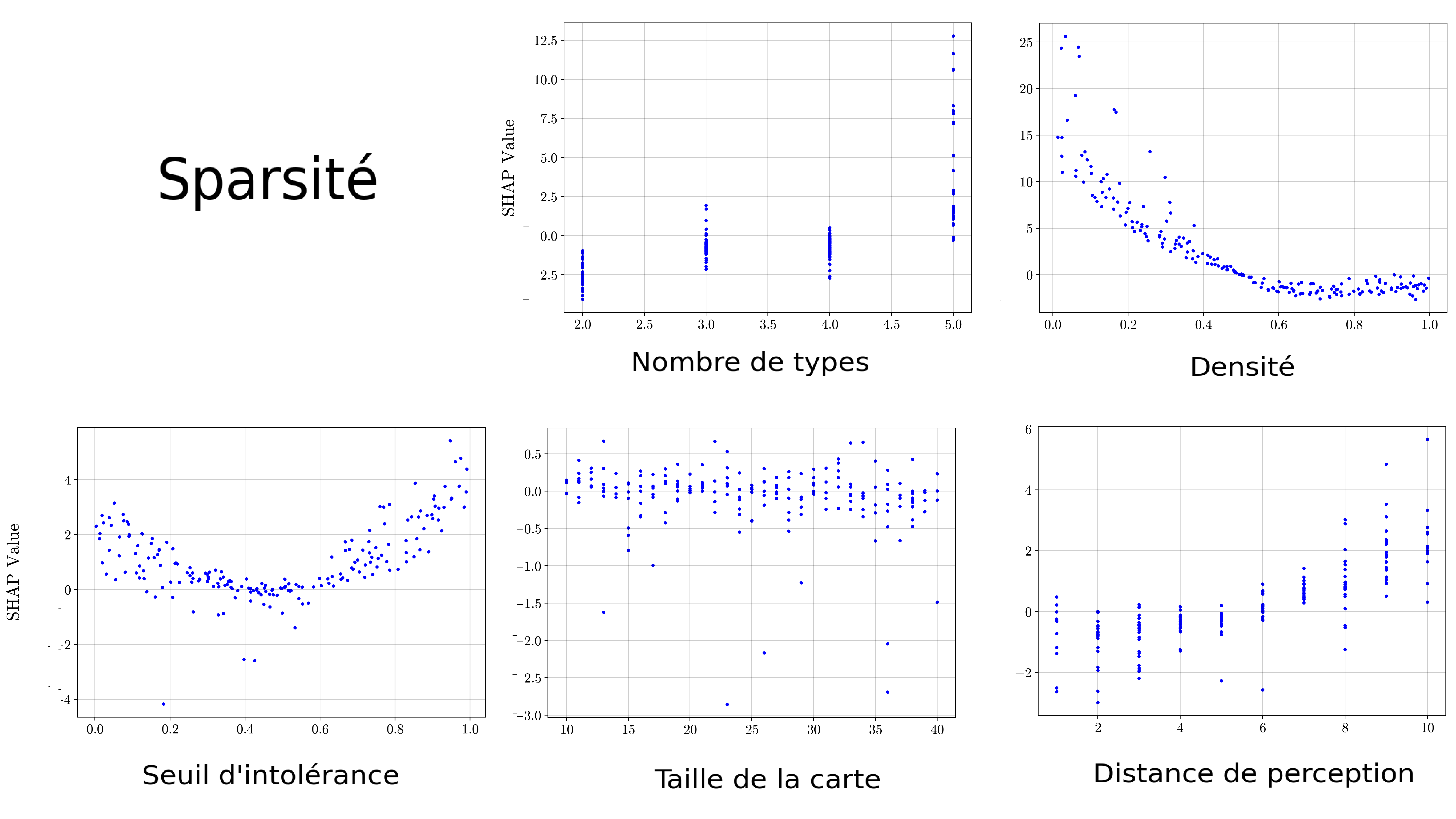}
\caption{Valeurs SHAP suivant les variables}
\label{fig:shap_vars}
\end{figure}

La Figure~\ref{fig:shap_vars} représente l'évolution des valeurs SHAP prédites par le modèle sur le jeu de données complet en fonction de chaque variable pour la convergence ainsi que pour la sparsité. Ces figures montrent notamment comment une variable impacte positivement ou négativement la prédiction suivant sa valeur. Notamment, la forme des nuages de points nous indique les relations linéaires, comme l'intolérance sur la non convergence et les relations plus complexes: la densité qui au-delà de 0.6 n'augmente plus la sparsité ou l'intolérance qui améliore la sparsité uniquement pour des valeurs faibles (équilibres non ségrégés) ou fortes (éparpillement du à la non convergence).

\begin{figure}[t]
\centering
\includegraphics[height=4.25cm,  width=0.47\textwidth]{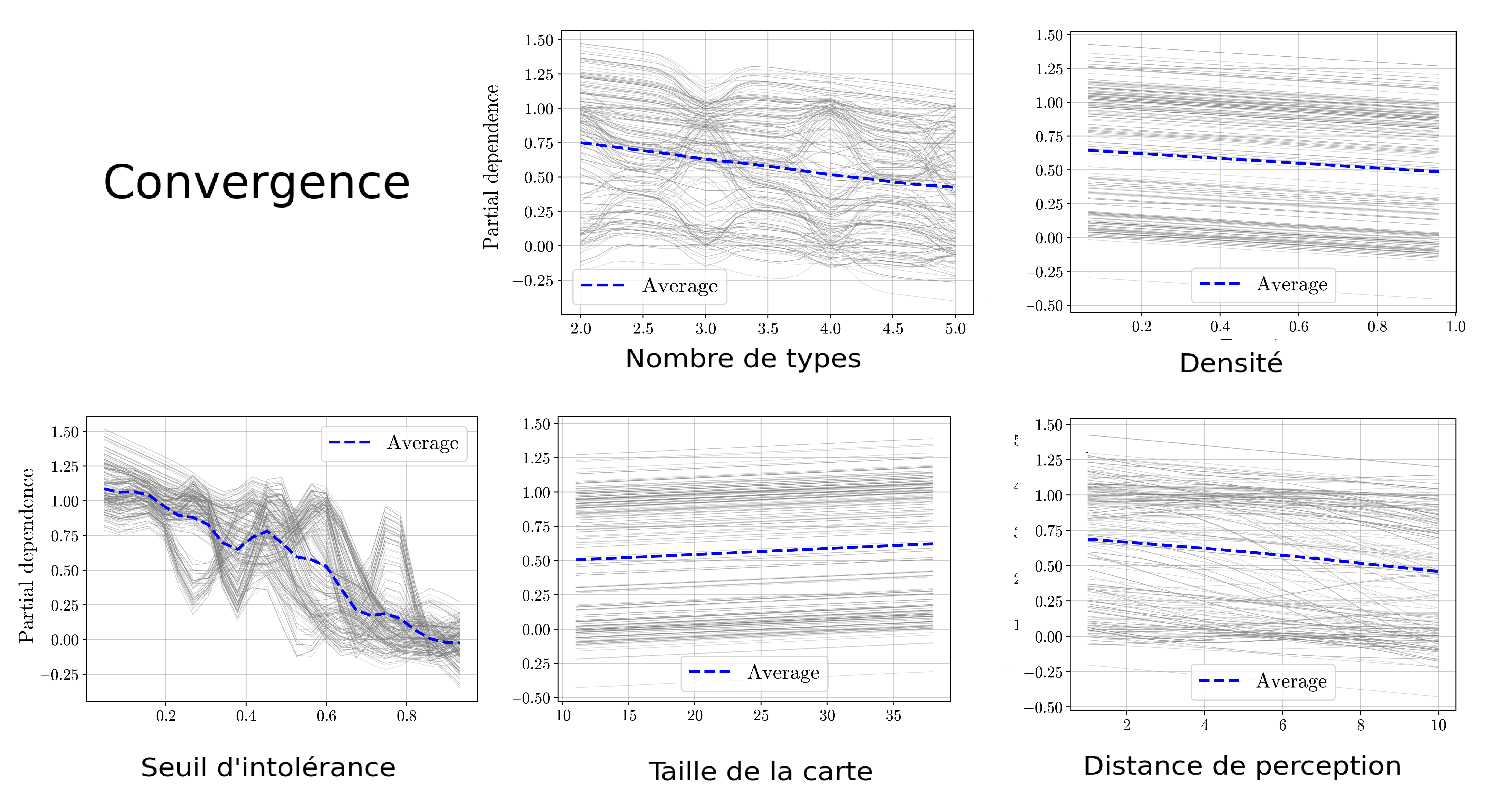}
\includegraphics[height=4.25cm,  width=0.47\textwidth]{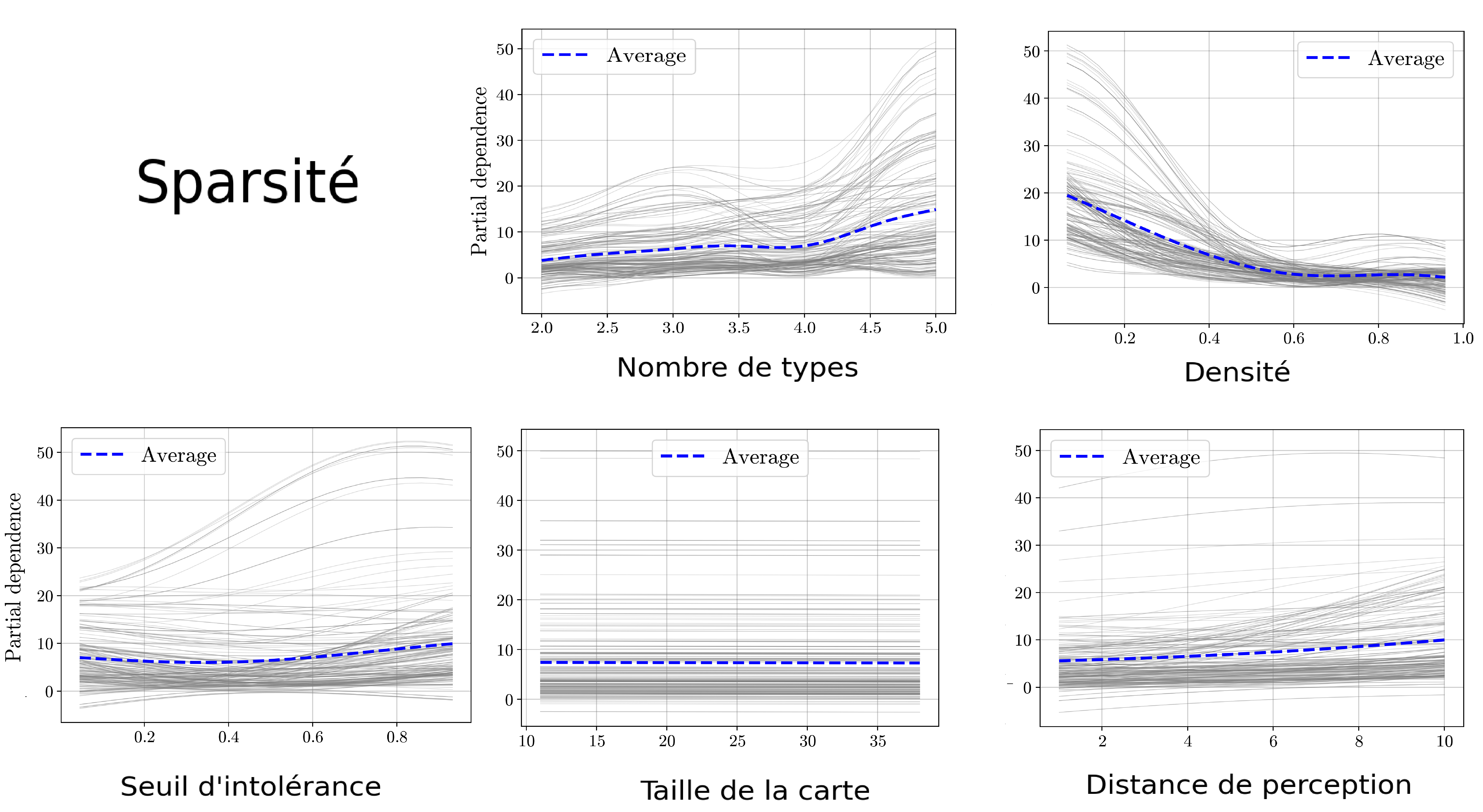}
\caption{Tracés PDP et ICE}
\label{fig:pdp_ice}
\end{figure}

Sur la Figure~\ref{fig:pdp_ice}, les Partial Dependence Plots (PDP) illustrent l’effet moyen de chaque variable sur la prédiction en neutralisant l’influence des autres, tandis que les courbes Individual Conditional Expectation (ICE) révèlent la variabilité des réponses individuelles, mettant en évidence des interactions entre les paramètres.  
Dans l’image présentée, les tendances globales observées dans les PDP/ICE – telles qu’un effet monotone ou non linéaire sur la sortie – sont confirmées par les dispersion des valeurs SHAP, qui révèlent également la présence d’interactions, bien que celles-ci soient relativement faibles pour certains paramètres. Ces analyses complémentaires offrent ainsi une compréhension approfondie et nuancée du comportement du modèle et de l’influence de chaque variable sur la simulation.
Les PDP montrent l’effet moyen de chaque variable sur la sortie prédite (par exemple, la ségrégation ou la sparsité), en neutralisant l’influence des autres paramètres. On peut ainsi repérer si l’effet d’une variable est monotone : plus la densité augmente, plus la convergence est difficile, ou s’il existe une relation non linéaire comme un optimum ou un point de bascule.  Les courbes ICE révèlent les trajectoires de prédiction pour chaque simulation individuelle. Si elles s’écartent sensiblement de la courbe PDP, cela met en évidence des interactions entre la variable étudiée et les autres variables ou facteurs aléatoires : le seuil d’intolérance peut augmenter ou diminuer la sparsité  suivant la distance de perception et suivant si la simulation converge ou non, d'où une moyenne PDP non représentative.
D'autres méthodes ou calculs sont égalements possibles à partir de nos modèles comme les indices de Sobol'~\cite{robani2025}, HSIC~\cite{iooss2022}, prédiction conforme~\cite{robani2025} ou LIME~\cite{blanco-volle2024} mais l'analyse de cette section a montré l'intérêt des modèle de substitution pour l'explicabilité des modèles à base d'agents et leurs adéquation avec le modèle malgré un apprentissage sur un jeu de données très parcimonieux.
\section{Conclusions et Perspectives}
\label{sec:conclu}
%
Dans ce papier, nous avons présenté une application des modèles de substitution pour l'intérprétabilité des simulations multi-agents. En particulier, en remplaçant les ABM par des modèles de substitution ayant appris sur seulement 50 configurations nous parvenons à reproduire fidèlement les comportements du système, tout en offrant une rapidité de calcul notable. Notre contribution vient du fait que nous ne validons pas seulement les modèles uniquement par rapport à leur capacité à prédire correctement mais également par rapport à leur capacité à expliquer des phénomènes globaux comme locaux, tendanciels comme émergents, et, pour ce faire, nous avons présenté comment des modèles rapides à évaluer pouvaient donner de nombreuses métriques d'interprétation des variables. 

Cependant, plusieurs défis subsistent, notamment en termes de robustesse, d'interprétabilité et de généralisation. Comment garantir la fiabilité de ces modèles sur des espaces de conception très hétérogènes ? Quelles stratégies adopter pour intégrer de manière optimale les incertitudes liées aux données et aux approximations du modèle de substitution ? Comment adapter nos modèles à la grande dimension ou aux grandes quantités de données ? Ces questions invitent à explorer des approches hybrides alliant intelligence artificielle, méthodes bayésiennes et techniques d'optimisation multi-fidélité. 
En particulier, il est essentiel d’exploiter la complémentarité des modèles de substitution en tenant compte de leurs forces et de leurs faiblesses. À cet égard, les approches ensemblistes apparaissent particulièrement adaptées, non seulement en raison de leur diversité, mais aussi parce qu’elles peuvent s’intégrer aussi bien à l’échelle macroscopique du modèle qu’à l’échelle microscopique des agents~\cite{blanco-volle2024}.
 
Par ailleurs, l'utilisation conjointe d'outils d'explicabilité — comme les PDP, ICE et les valeurs SHAP — offre une double perspective : une vision globale des tendances et une analyse fine des interactions locales entre variables. Ces analyses ouvrent la voie à de nombreuses pistes de recherche, telles que l'intégration en temps réel de modèles de substitution dans des boucles de simulation interactive, l'adaptation de méthodes explicables aux modèles hybrides (combinant variables continues et catégoriques) ou encore l'application de ces outils pour l'extrapolation des séries temporelles issues de simulations dynamiques. En somme, les avancées en IA et en modélisation hybride pourraient s'intégrer aux ABM dans des domaines variés en offrant des solutions à la fois efficaces, interprétables et adaptées aux exigences des applications de grande échelle~\cite{iooss2022}.

D'autres extensions possible concernent les approches séquentielles comme l'optimisation Bayésienne multi-objectif pour enrichir progressivement le plan d'expérience afin d'améliorer les modèles tout en maîtrisant le budget computationnel ce qui pose la question de l'évaluation des performance des modèle qui est rarement simple en pratique, car l'ABM ne renvoie que des métriques agrégées qui ne rendent pas forcément compte des comportements émergents que le modèle pourrait chercher à explorer. Les mêmes questions se posent pour un objectif de calibration de paramètres grâce aux modèles. On voudrait ainsi  pouvoir quantifier à quel moment un modèle a suffisamment appris et à quels moment ses prédictions sont de confiance. 

\textbf{Remerciements.} Les auteurs remercient le projet de recherche MIMICO financé par l'Agence Nationale de la Recherche (ANR)  n$^o$ ANR-24-CE23-0380. 
Nous sommes reconnaissant à Elisa Cueille et Matthieu    Mastio pour leur aide avec GAMA.

\end{document}